\def\year{2021}\relax
\documentclass[letterpaper]{article} 
\usepackage{aaai21}  
\usepackage{times}  
\usepackage{helvet} 
\usepackage{courier}  
\usepackage[hyphens]{url}  
\usepackage{graphicx} 
\usepackage{graphicx}  
\usepackage{natbib}  
\usepackage{caption} 
\usepackage{xspace}
\usepackage{amsmath}
\usepackage{graphics}
\usepackage{epsfig}
\usepackage{amssymb}
\usepackage{color}

\usepackage{algorithm}
\usepackage{algpseudocode}
\usepackage{bm}
\usepackage[switch]{lineno}

\title{Object-aware Feature Aggregation for Video Object Detection}
\author{
Qichuan Geng$^1$\hspace{0.5cm}Hong Zhang\hspace{0.5cm}Na Jiang$^2$\hspace{0.5cm}Xiaojuan Qi$^{3}$\hspace{0.5cm}Liangjun Zhang$^{4}$\hspace{0.5cm}Zhong Zhou$^1$
\vspace{0.2cm}\\  
{$^1$Beihang University \quad $^2$Capital Normal University \\ \quad $^3$ The University of Hong Kong \quad $^4$ Baidu, Inc.}
\vspace{0.2cm} \\
{\tt\small zhaokefirst@buaa.edu.cn \quad \{fykalviny, zhouzhong2011\}@gmail.com}
\vspace{-0.01in}
{\tt\small \quad jiangna@cnu.edu.cn \quad xjqi@eee.hku.hk \quad liangjunzhang@baidu.com}
}

\newcommand{\namefull}{Object-aware Feature Aggregation}
\newcommand{\name}{OFA}
\newcommand{\modulename}{Object-aware Feature Aggregation}
\newcommand{\modulenameshort}{OFA}
\newcommand{\modulepatha}{semantic path}
\newcommand{\modulepathb}{localization path}

\makeatletter
\DeclareRobustCommand\onedot{\futurelet\@let@token\@onedot}
\def\@onedot{\ifx\@let@token.\else.\null\fi\xspace}
  
\def\ie{\emph{i.e}\onedot}

\makeatother

\begin{document}
\maketitle

\begin{abstract}
We present an \modulename{}~(\modulenameshort{}) module for video object detection~(VID). Our approach is motivated by the intriguing property that video-level object-aware knowledge can be employed as a powerful semantic prior to help object recognition. 
As a consequence, augmenting features with such prior knowledge can effectively improve the classification and localization performance.
To make features get access to more content about the whole video, we first capture the object-aware knowledge of proposals and incorporate such knowledge with the well-established pair-wise contexts.
With extensive experimental results on the ImageNet VID dataset, our approach demonstrates the effectiveness of object-aware knowledge with the superior performance of $83.93\%$ and $86.09\%$ mAP with ResNet-101 and ResNeXt-101, respectively. When further equipped with Sequence DIoU NMS, we obtain the best-reported mAP of $85.07\%$ and $86.88\%$ upon the paper submitted. The code to reproduce our results will be released after acceptance.
\end{abstract}
\section{Introduction}
\label{Introduction}
Convolutional neural networks~(CNNs)~\cite{krizhevsky2012imagenet, simonyan2014very, szegedy2015going, he2016deep, huang2017densely, cao2019gcnet, li2018recurrent, li2018unified, pan2016learning, qiu2017learning, simonyan2014two} have gained remarkable success in image object detection~\cite{girshick2014rich,ren2015faster,law2018cornernet,zhou2019objects,liu2016ssd,fu2017dssd}. Video object detection~(VID) extends the idea of localizing and recognizing objects in videos.  
But beyond the single image, video object detection is a more challenging task that suffers from difficulties like motion blur, out-of-focus and occlusion.

Videos contain various spatial-temporal cues which can be exploited to overcome the aforementioned problems.
Several studies have been carried out to improve the performance of localizing and recognizing with spatial-temporal context.
Box-level association methods~\cite{feichtenhofer2017detect,han2016seq,kang2016object,kang2017object,kang2017t} link bounding boxes of each frame according to different constraints. 
Spatio-temporal feature aggregations~\cite{wang2018fully,xiao2018video,zhu2018towards,zhu2017flow,wu2019sequence,deng2019relation,chen2020memory} augment features by capturing contexts across adjacent frames. 
Recent works~\cite{wu2019sequence,deng2019relation,chen2020memory} have shown that attention-based operators can generate powerful features to obtain more accurate and stable predictions. As the attention operators are capable of reducing the variance within features via extracting highly semantically similar pair-wise contexts from the whole sequence. 

\begin{figure}[t]
    \centering
    \includegraphics[width=\linewidth]{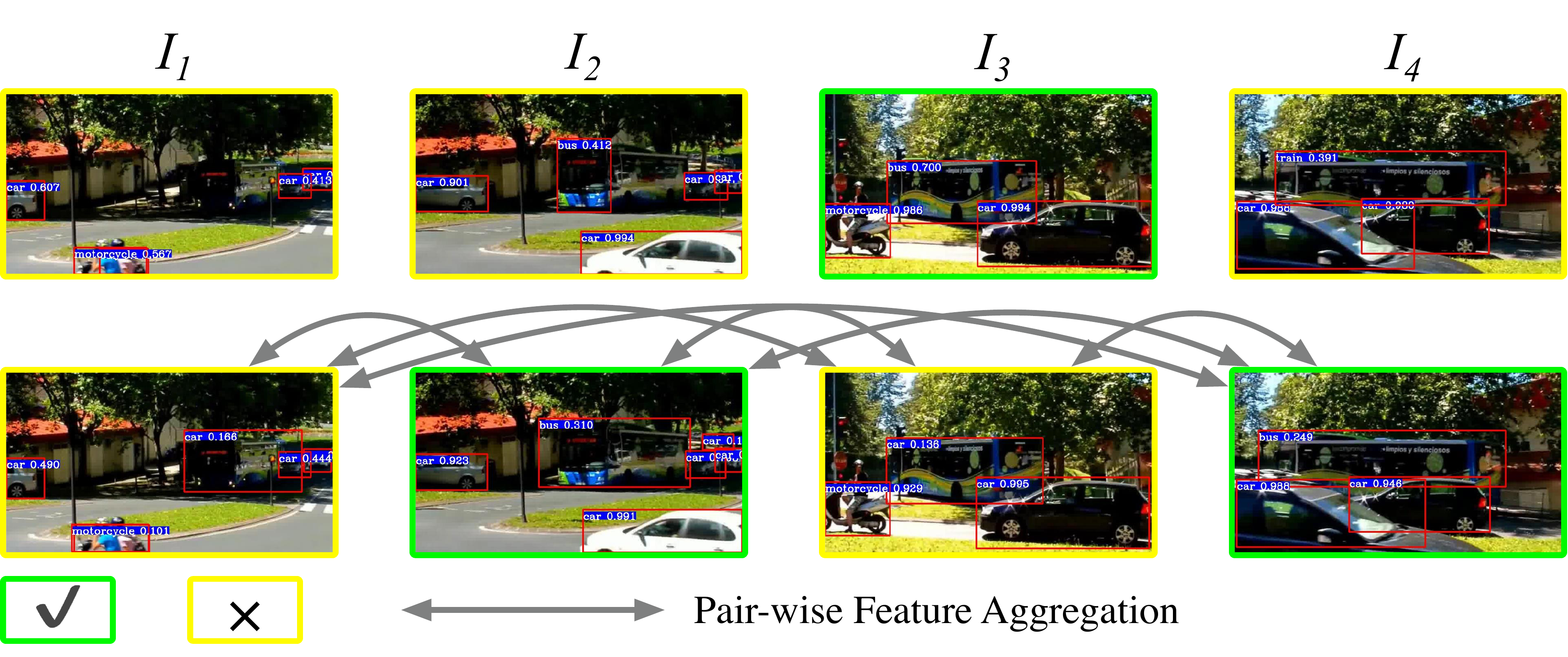}
    \caption{
        \textbf{Example outputs of the previous state-of-the-art attention-based approach}~\cite{wu2019sequence}. The first row is the result without temporal aggregation, which is produced by the single frame version, and the second row is the result which aggregates pair-wise context from support frames. 
    }
    \label{fig:analysis}
\end{figure}

Existing works modeling the spatial-temporal contexts mainly focus on short-time windows or pair-wise feature relations.
In comparison, less effort has been made to incorporate global video content. 
Fig.~\ref{fig:analysis} visualizes several frames containing the \emph{bus} category. Obviously, after passing the context information from other proposals, the recognition ability of \emph{bus} is increasing. However, as shown in Fig.~\ref{fig:analysis}($I_1$), there is a limited improvement when the appearance of the object varies substantially. Moreover, the prediction of the salient object will be interfered by other proposals~(\emph{bus} is wrongly categorized as \emph{car} after the propagation in Fig.~\ref{fig:analysis}($I_3$)). 

To mitigate the situation above, we revisit the mechanism of humans to recognize videos.
There is an intriguing observation that when people are not certain about the identity of objects with deteriorated appearances in the reference frames, it is natural to seek the existence of objects in the video and assign them with the most similar ones. As the prior provides the video-level constraints of object appearance and category, we refer to it as the \emph{object-aware knowledge}. 
It narrows down the scope of categories to be assigned due to the capability of capturing the global content of the video.
Meanwhile, the object-aware knowledge about appearances serves as references to facilitate the bounding box predictions.

In this paper, we propose an \modulename{}~(OFA) module to distill this insight into video object detection. 
As the \emph{object-aware knowledge} is usually extracted from isolated regions containing objects, it is hard to obtain such information before being detected. Therefore, our \modulenameshort{} module selectively aggregates features corresponding to proposals. By incorporating \emph{object-aware knowledge} with the well-established pair-wise features, we can effectively improve the performance of video object detection.
Concretely, the input features are split into two parts and fed into separate paths, \ie, \emph{semantic path} and \emph{localization path}.
a) \textbf{Semantic path.} We first collect the object-aware knowledge from proposals across the whole video. Meanwhile, the pair-wise semantic context is obtained via calculating the similarity between proposals. By aggregating above pair-wise semantic context and object-aware knowledge of proposals, the features encode more knowledge about other regions and the whole video.
b) \textbf{Localization path.} Localization features are also augmented with the pair-wise localization context. In contrast to the semantic path, we locally enhance the features to ensure the features sensitive to relative positions.

To further improve the performance, we propose an efficient post-processing strategy named Sequence DIoU NMS. Instead of linking the bounding boxes according to the intersection of union~(IoU), we perform Distance-IoU~(DIoU)~\cite{zheng2020distance} to associate and suppress boxes in videos.

Though being simple, the \modulenameshort{} module achieves surprisingly good performance, and it is easy to complement existing attention-based and proposal-based video object detection methods. 
Extensive empirical evaluations explicitly demonstrate that the proposed method is competitive in performance. Furthermore, we conduct ablation experiments to analyze the effects of various design choices in the \modulenameshort{} module.

\section{Related Work}
\label{Related Work}

\begin{figure*}[t]
    \centering
    \includegraphics[width=\linewidth]{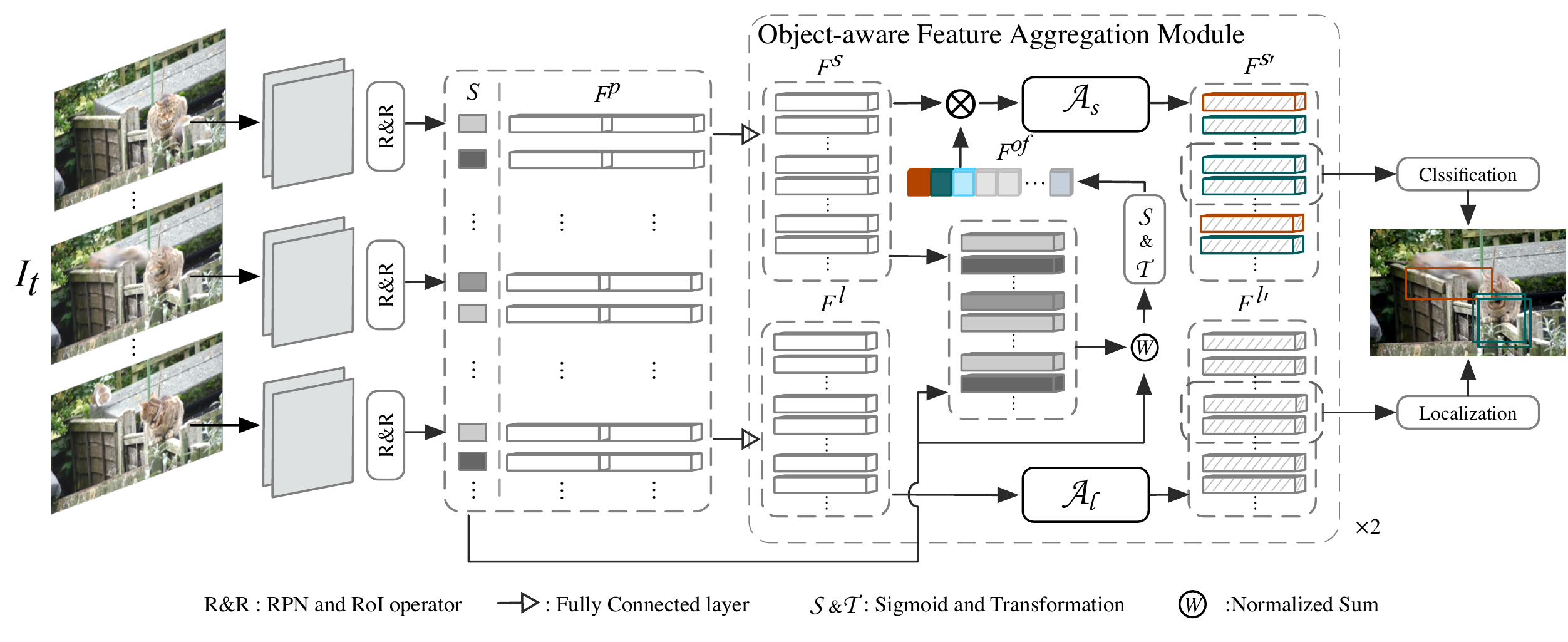}
    \caption{\textbf{Overview of the \modulename{} module for video object detection}. Given the input reference frame $I_{t}$ and support frames, their objectness score $S = \{s^{p}\}$ and features $F^p = \{f^{p}\}$ corresponding to object proposals are produced by the Region Proposal Network~{(RPN)} and the succeeding RoI operator. We split $f^{p}$ into $f^{s}$ and $f^{l}$, which are augmented in two parallel paths of the \modulenameshort{} module. 
    The semantic features and localization features of the reference frame are augmented by its relation features over all the $f^{s}$ and $f^{l}$. To aggregate all proposals corresponding to the main body or salient parts of the objects, we propose the effective and efficient \emph{Object-aware Knowledge Extraction} to highlight class-dependent channels of semantic features. With the help of object-aware knowledge, the aggregated semantic features tend to be consistent with the whole global context.}
    \label{fig:overview}
\end{figure*}

\paragraph{Image Object Detection.} Thanks to the advances in deep neural networks~\cite{he2016deep,krizhevsky2012imagenet,simonyan2014very,szegedy2015going} and large-scale annotated datasets~\cite{russakovsky2015imagenet,lin2014microsoft}, several state-of-the-art image object detection approaches~\cite{girshick2014rich,girshick2015fast,ren2015faster,he2015spatial,dai2016r,law2018cornernet,lin2017feature,lin2017focal,liu2016ssd,redmon2016you,redmon2017yolo9000,redmon2018yolov3,fu2017dssd} are proposed. The proposal-based approaches~\cite{girshick2014rich,girshick2015fast,ren2015faster,dai2016r,lin2017feature} include the stages of feature extraction, proposal generation and bounding box prediction. 
Region Proposal Network~(RPN)~\cite{ren2015faster} was proposed to generate proposals with the assistance of predefined anchors. Beyond Spatial pyramid pooling~\cite{he2015spatial}, RoIPooling~\cite{girshick2015fast}, RoIAlign~\cite{he2017mask}, position-sensitive RoIPooling~\cite{dai2016r} and other similar operations were introduced to extract features from proposals and improved the efficiency. Inspired by the attention mechanism~\cite{vaswani2017attention}, Relation Network~\cite{hu2018relation} proposed an object relation module to model object relations in a single image. With detailed experiments~\cite{wu2020rethinking}, it was proved that different head structures had opposite preferences towards classification and localization.

Although anchors and proposals have boosted the accuracy of detectors, it also hurts the efficiency of the whole network. Hence, several recent works are dedicated to achieving a real-time detection speed under the premise of high precision.
One-stage proposal-free object detectors~\cite{lin2017focal,liu2016ssd,redmon2016you,redmon2017yolo9000,redmon2018yolov3,fu2017dssd} and anchor-free methods~\cite{law2018cornernet,zhou2019objects} recognized and localized objects on the feature map extracted by the CNN backbone. These approaches deeply analyzed the mechanism of CNN-based detectors and pointed out new directions for improving detectors. However, to study the effect of the video-level object-aware knowledge of proposals on video object detection, we follow the well-studied two-stage detection pipeline.

\paragraph{Video Object Detection.} To extend the image object detectors to the video domain, two main directions are proposed to leverage the spatio-temporal information. The first direction is to associate related boxes in time and space.
To incorporate motion information, D\&T~\cite{feichtenhofer2017detect} integrated tracking and detection in a unified framework to promote each other. Tubelet~\cite{kang2016object,kang2017object,kang2017t} was built on anchor cuboids in consecutive frames to ensure temporal continuous. As a dense motion representation in frames, optical flow~\cite{wang2018fully,zhu2017deep,zhu2017flow} can be used to align and warp features extracted from adjacent frames to boost video object detection, which heavily relied on accurate motion estimation. Seq-NMS~\cite{han2016seq} made temporal links according to Jaccard overlap between bounding boxes of adjacent frames and rescored predictions in the optimal paths. 

The second direction is to enhance features by aggregating related contexts from frames. 
STSN~\cite{bertasius2018object} directly predicted sampling locations in different frames without relying on optical flow.
STMN~\cite{xiao2018video} proposed Spatial-Temporal Memory Module to model long-term appearance and movement changes. 
Beyond previous methods utilizing features from a short temporal window, SELSA~\cite{wu2019sequence} considered the semantic impact between related proposals in all the frames. RDN~\cite{deng2019relation} distilled relation through repeatedly refining supportive object proposals with high confidences, which were used to upgrade features. MEGA~\cite{chen2020memory} considered both local and global aggregation to enhance the feature representation, which overcame the ineffective and insufficient problem.
The above methods verified the merit on pair-wise object relation to eventually enhance feature discrimination.

Consistent with human cognitive behavior, including 
the prior relevant to objects in the entire video can directly reduce the difficulty of object recognition and localization. So different from the above methods, we directly extract the object-aware knowledge of proposals in the video and try to provide the network with the knowledge about what objects are in these video frames.
\section{Method}
\label{method}
In this section, we elaborate on how we devise the \modulename{}~(OFA) module to enable the whole architecture to fully utilize object-aware knowledge for detection in videos.
Incorporating object-aware knowledge with pair-wise contexts, we integrate the \modulenameshort{} module to improve feature discrimination. Therefore, guided by the video-level object-aware knowledge, the aggregated features tend to be consistent with this whole global context.

Roughly, given the reference frame and support frames, proposals are produced with the faster-rcnn-style manner. Then, the proposal features $F^{p} = \{f^{p}_1, f^{p}_2, ...\}$ are enhanced with two stacked \name{} modules. 
Two independent paths, \ie, \textit{semantic path} and \textit{localization path}, are designed to aggregate the object-aware features with the semantic and localization contexts for classification and localization. Sequence DIoU NMS is introduced to further improve the performance in the post-processing stage. 
Fig.~\ref{fig:overview} gives an overview of the proposed approach. 
\subsection{Pair-wise Feature Aggregation}
The basic idea of pair-wise feature aggregation~\cite{wu2019sequence,deng2019relation,chen2020memory} is to measure relation features of one object as the weighted sum of features from other proposals, and the weights reflect object dependency in terms of both semantic and/or localization information. Previous works have demonstrated that spatial-temporal feature aggregations help to enhance the performance of detectors. 
Formally, suppose that we are aggregating the features $F = \{f_i\}$, the learnable association between proposals serves as weights to aggregate features from others as follows: 

\begin{equation}
    \begin{aligned}
        \delta_i &= softmax(\frac{\phi(f_i) \psi(F)^T}{\sqrt d}) \\
        \mathcal{A}(f_i, F) &= \text{ReLU}(f_i + \delta_i \cdot F), \\
    \label{eq:pw}
    \end{aligned}
\end{equation}
where both $\phi$ and $\psi$ are linear learnable transformation functions. $d$ is the dimension of $f_i$, where $i$ is the index of proposals. 

\subsection{\namefull{} Module}
As aforementioned, pair-wise context aggregations of proposals help alleviate the problem of appearance degradation in video object detection. However, such knowledge is insufficient for the proposals which are quite different from others.
Intuitively, knowledge in the proposal feature space corresponds to meaningful prior information about existing objects, thus we obtain the rich object-aware knowledge by aggregating the global context of proposals, which is able to serve as a semantic prior of the video to recognize objects. Nevertheless, proposals contain false positive samples which include only parts of objects. The information cannot provide an accurate global context.

\paragraph{Object-aware Knowledge Extraction.} The \name{} module is designed to emphasize on the high-quality proposals. Obviously, the proposals with high objectness scores are most likely to contain the main area or salient parts of the objects. Moreover, each frame consists of clusters of object proposals, which help eliminate the uncertainty of objectness scores produced by RPN. For a specific sequence $\{f_i\}$ with objectness score $\{s_i\}$, we design the object-aware features extractor $G_{of}$, which votes the proposals with the corresponding objectness scores. Additionally, to maintain the magnitude of features, we normalize the features across all proposal scores. The object-aware feature is obtained as:



\begin{equation}
    G_{of}(F) = \frac{\sum_i(s_i \times f_i)}{\sum_i(s_i)}.
    \label{eq:hgc}
\end{equation}


Considering the two paths in our framework, proposal features $F^{p}$ are split into $F^{s} = \{f^{s}_1, f^{s}_2, ...\}$ and $F^{l} = \{f^{l}_1, f^{l}_2, ...\}$ which are fed into the \emph{semantic path} and the \emph{localization path} respectively. We obtain the aggregated features $F^{s}{'} = \{f^s_1{'} , f^s_2{'} , ...\}$ and $F^{l}{'} = \{f^l_1{'} , f^l_2{'} , ...\}$ as follows:
\vspace{-0.5cm}
\paragraph{Semantic Path: Aggregating the object-aware features with the semantic context.}
To generate object-aware features ${F^{of}}$, we first estimate a set of factors from Eqn.~\ref{eq:hgc} to selectively highlight the class-dependent channels of $F^s$. Then we employ the pair-wise semantic features aggregation $\{\mathcal{A}_s(\cdot, \cdot)\}$ with Eqn.~\ref{eq:pw}. Combining with the object-aware feature, we have
\begin{equation}
    \begin{aligned}
        F^{of} &= \{softmax(\mathcal{T}(G_{of}(F^{s})))\otimes f^s_i\} \\
        f^s_i{'} &= \mathcal{A}_s(f^{of}, F^{of}) ,
    \label{eq:cls}
    \end{aligned}
\end{equation}
where $\mathcal{T}(\cdot)$ is non-linear transformations applied to the object-aware feature. 
\vspace{-0.5cm}
\paragraph{Localization Path: Aggregating the localization context.}
As localization contexts among proposals usually focus on short temporal windows and the object-aware knowledge may harm the overall performance, we design a parallel localization branch urging the relation module to focus only on localization features.
To keep sensitive to relative position, $F^l$ is aggregated as follows:
\begin{equation}
    f^l_i{'} = \mathcal{A}_l(f^l_i, F^l).
    \label{eq:loc}
\end{equation}
To avoid being interfered by irrelevant semantic-similar contexts, we get the augmented feature $\{f^l_i{'}\}$ by integrating localization contexts with individual learnable $\phi_l$ and $\psi_l$ functions. 

\subsection{Sequence DIoU NMS}
The predictions from faster-rcnn-style detectors consist of a series of redundant results that need to be filtered. Considering temporal consistency, sequence post-processing methods link detection boxes across consecutive frames and select optimal tubelets to suppress redundant predictions. 
Seq-NMS~\cite{han2016seq} links the bounding boxes according to intersection over union~(IoU). 
With modified criteria~\cite{deng2019relation, gkioxari2015finding}, smooth tubelets could be selected.
As objects are usually dense and occlude each other in consecutive frames, the criterion to decide whether different bounding boxes correspond to the same object is a key and difficult factor.

Without extra inputs like pair-wise relation weight, we employ the Distance-IoU~(DIoU)~\cite{zheng2020distance} to link and suppress boxes, which is an extended overlap metric in the image object detection to distinguish crowded objects. 
With the center distance of bounding boxes $\rho(\cdot, \cdot)$ and diagonal length of the union box of inputs $c(\cdot, \cdot)$, we have

\begin{equation}
    \begin{aligned}
        \text{DIoU}(\cdot, \cdot) = \text{IoU}(\cdot, \cdot) - \frac{\rho^2(\cdot, \cdot)}{c^2(\cdot, \cdot)}
    \end{aligned}
    \label{eq:diou}    
\end{equation}

In our proposed Sequence DIoU NMS, we recursively maximize the sum of object scores subject the modified constraint to find maximum paths $B^{seq} = \{b_{t_0}[i_{t_0}],..., b_{t_1}[i_{t_1}]\}$ as follows:
\begin{equation}
    \begin{aligned}
        \textbf{\textit{i}}' &= \mathop{argmax}\limits_{i_{t_0},..., i_{t_1}} \sum_{t=t_0}^{t_1}c_t[i_t] \\
        C^{seq} &\leftarrow avg(C^{seq}) \\
        & s.t. \\
        & 1 \leq t_0 \leq t_1 \leq T, \\
        & \text{DIoU}(b_t[i_t], b_{t+1}[i_{t+1}]) \geq \tau_1,\\
    \end{aligned}
    \label{eq:seqdnms}
\end{equation}

The optimization returns a set of indices $\textbf{\textit{i}}'$ to rescore the detection confidence $C^{seq} = \{c_{t_0}[i_{t_0}],..., c_{t_1}[i_{t_1}]\}$, with its average value.

Although being simple, Sequence DIoU NMS further improves the performance in our experiments. We present the pseudo-code of Sequence DIoU NMS in Algorithm~\ref{alg:seq-dnms}.

\begin{algorithm}[h] 
    \caption{Sequence DIoU NMS.} 
    \label{alg:seq-dnms} 
    \begin{algorithmic}[1] 
        \State {\bfseries Input:} ${\{\bm b_t\}}$, ${\{\bm c_t\}}$, $\tau_1$, $\tau_2$,
        \State $t \in \{1,...,T\}$ is the time stamp,
        \State ${\bm b_t=\{b_t[i]\}}$ is the list of initial bounding boxes in frame $t$, where $i$ is the indice,
        \State ${\bm c_t=\{c_t[i]\}}$ contains corresponding detection confidence,
        \State $\tau_1$ and $\tau_2$ are the thresholds.
        \State {\bfseries Ensure:} optimal $\{ \bm b_t\}$, $\{\bm c_t\}$.
        \State Create linkes $L$:
        \For{$t=1$ {\bfseries to } $T-1$}
        \State $L_t = \O$;
            \For{$b_t[i] \in {\bm b_t}$ , $b_{t+1}[j] \in {\bm b_{t+1}}$}
                \If{$DIoU(b_t[i], b_{t+1}[j])\ge \tau_1$}
                    \State $L_t \cup link(b_t[i], b_{t+1}[j])$;
                \EndIf
            \EndFor
        \State $L = L \cup L_t$;
        \EndFor

        \State Find maximum paths: 
        \While{$L \neq \O$}
        \State find the maximum path $B^{seq}$ from $t_0$ to $t_1$;
            \If{$B^{seq} = \O $} 
                \State stop loop;
            \EndIf
            \For{$t=t_0$ {\bfseries to } $t_1$}
                \For{$b_t[k] \in \bm b_t$}
                    \If{$DIoU(b_t[i_t], b_t[k])\le \tau_2$}
                    \State delete $b_t[k]$ and corresonding links;
                    \EndIf
                \EndFor
            \EndFor
        \State update $L$;
        \State rescore $C^{seq}$ according to Eqn.~\ref{eq:seqdnms};
        \EndWhile
    \State {\bfseries Output:} The rescored result of the sequences: ${\{\bm c_t\}}$.
    \end{algorithmic} 
\end{algorithm}

\subsection{Relation to Other Approaches}

\paragraph{Global features aggregation.} The global context has gained success on a variety of tasks like semantic segmentation~\cite{liu2015parsenet,zhao2017pyramid}, action recognition~\cite{wang2018videos} and similar tasks.
However, most of the previous works~\cite{liu2015parsenet,zhao2017pyramid,zhang2018context} capture global context from both foreground and background, without concentrating on foreground objects. 
In a two-stage video object detection pipeline, proposals usually contain rich foreground knowledge for object classification. This motivates us to adopt the global context operator in detection. Although the global context of proposals is employed in the action recognition~\cite{wang2018videos}, however, it does not emphasize those proposals containing salient views of objects. 
Our approach is the first to extract object-aware knowledge from object proposals to improve the performance in VID. Remarkably, with prior knowledge, we design a novel feature aggregation strategy to avoid the negative effect of redundant or low-quality proposals.
\vspace{-2ex}
\paragraph{Multi-head attention-based approaches.} Existing multi-head attention works~\cite{vaswani2017attention,deng2019relation} are designed with multiple parallel attention operators to augment features. They capture enriched relations by collecting the features from different heads. In fact, video object detection is to classify and regress objects, hinting that the separable branches for classification and regression may boost the detection performance. Therefore, our approach divides features into two independent branches and optimizes with different supervision. Importantly, this is a necessary step to augment the semantic features and keep the localization features spatial sensitive.
\section{Experiments}
\label{Experiments}

\begin{table*}[ht]
    \begin{center}
    \begin{tabular}{c | c | c}
        \hline
            ~Methods~&~Backbone~&~mAP~ \\
        \hline
            FGFA~\cite{zhu2017flow}~&~ResNet-101~&~76.30~\\
            D\&T~\cite{feichtenhofer2017detect}~&~ResNet-101~&~75.80~\\
            MANet~\cite{wang2018fully}~&~ResNet-101~&~78.10~\\
            SELSA~\cite{wu2019sequence}~&~ResNet-101~&~82.69~\\
            RDN~\cite{deng2019relation}~&~ResNet-101~&~81.80~\\
            MEGA~\cite{chen2020memory}~&~ResNet-101~&~82.90~\\
            FEVOD~\cite{jiang2019learning}~&~ResNet-101~&~80.10~\\
            \textbf{Ours}~&~\textbf{ResNet-101}~&~\textbf{83.93}~\\
        \hline
            FGFA*~\cite{zhu2017flow}~&~ResNet-101~&~78.40~\\
            MANet*~\cite{wang2018fully}~&~ResNet-101~&~80.30~\\
            ST-Lattice*~\cite{chen2018optimizing}~&~ResNet-101~&~79.60~\\
            D\&T*~\cite{feichtenhofer2017detect}~&~ResNet-101~&~79.80~\\
            STMN*+~\cite{xiao2018video}~&~ResNet-101~&~80.50~\\
            RDN*~\cite{deng2019relation}~&~ResNet-101~&~83.8~\\
            MEGA*~\cite{chen2020memory}~&~ResNet-101~&~84.5~\\
            FEVOD*~\cite{jiang2019learning}~&~ResNet-101~&~82.10~\\
            \textbf{Ours*}~&~\textbf{ResNet-101}~&~\textbf{85.07}~\\
        \hline
            D\&T*~\cite{feichtenhofer2017detect}~&~ResNeXt-101~&~81.60~\\
            SELSA~\cite{wu2019sequence}~&~ResNeXt-101~&~84.30~\\
            RDN~\cite{deng2019relation}~&~ResNeXt-101~&~83.2~\\
            RDN*~\cite{deng2019relation}~&~ResNeXt-101~&~84.7~\\
            MEGA~\cite{chen2020memory}~&~ResNeXt-101~&~84.1~\\
            MEGA*~\cite{chen2020memory}~&~ResNeXt-101~&~85.4~\\
            \textbf{Ours}~&~\textbf{ResNeXt-101}~&~\textbf{86.09}~\\
            \textbf{Ours*}~&~\textbf{ResNeXt-101}~&~\textbf{86.88}~\\
         \hline
    \end{tabular}
    \end{center}
    \caption{Performance comparison with state-of-the-art systems on the ImageNet VID validation set. $+$ indicates the use of model emsembling. $*$ indicates the use of sequence post-processing methods (e.g Seq-NMS, tubelet rescoring, and our Sequence DIoU NMS). }
    \label{tab:performance}
\end{table*}

\subsection{Dataset and Evaluation}

We evaluate our method on the ImageNet VID dataset~\cite{russakovsky2015imagenet}, which is a large scale benchmark for the video object detection task. The ImageNet VID dataset consists of 3862 training videos and 555 validation videos. There are 30 object categories annotated in this dataset. Meanwhile, we follow the common protocols~\cite{wu2019sequence, deng2019relation, chen2020memory} to mix the ImageNet VID dataset with the Image DET dataset for training. We evaluate our method on the validation set and use mean average precision (mAP) as the main evaluation metric. Furthermore, we also report motion-specific mAP on the validation set to illustrate the effectiveness of our approach.

\subsection{Network Architecture}
\paragraph{Backbone.} We use ResNet-101~\cite{he2016deep} and ResNeXt-101~\cite{xie2017aggregated} as backbone networks. For those two backbone networks, we enlarge the resolution of feature maps in the last stage by halving the strides and doubling the dilation rates of convolutions. We make ablation experiments mainly with ResNet-101. For the more powerful ResNeXt-101, we report the final results.

\paragraph{Region Feature Extraction.} We apply RPN on the top of the \emph{conv4} stage. In RPN, the anchors of 3 aspect ratios $\{1:2, 1:1, 2:1\}$ and 3 scales $\{128^2, 256^2, 512^2\}$ are predefined on each spatial location for proposal generation. With generated proposals, we apply RoIAlign~\cite{he2017mask} on the \emph{conv5} stage and a 1024-D fully-connected~(FC) layer to extract $F^p$ for proposals.

\paragraph{\modulenameshort{} Module.} We split the 1024-D $f^{p}_i$ into two 512-D $f_i^{s}$ and $f_i^{l}$. In the \modulepatha{} and \modulepathb{}, the internal channels of pair-wise feature aggregation are 512 after inserting a FC layer with 512 channels. 
In the stage of extracting object-aware knowledge of proposals, the $\mathcal{T}(\cdot)$ is implemented with FC-ReLU-FC with 512 channels.

\paragraph{Sequence DIoU NMS.} We conduct our Sequence DIoU NMS by slightly modifying the original Seq-NMS. In contrast to other methods, our Sequence DIoU NMS does not need extra inputs. Without bells and whistles, the network predictions can be improved effectively with the setting of $\tau_1=0.6$ and $\tau_2=0.5$.

\subsection{Implementation Details}

Our approach is mainly built on SELSA~\cite{wu2019sequence}. The input frames are resized to a shorter side of 600 pixels. The network is trained on 8 Nvidia P40 GPUs. A total of 6 epochs of SGD training is performed with a batch size of 8, with a learning rate of $2.5 \time 10^{-4}$ and $2.5 \time 10^{-5}$ in the first 4 epochs and in the last 2 epochs, respectively. In the training phase, three frames are sampled from the same given video. Except photometric distortion, we apply the same data augmentation, including random expand, crop and flipping, to these frames to keep them aligned. In the test phase, 21 frames from the given video are sampled.

\paragraph{Training our \modulenameshort{} module.} In training and test phases, different strategies are applied to generate proposals. To ensure consistency in both two phases, we use the class-agnostic objectness scores provided by RPN in these two phases. Furthermore, we block the gradients propagated from the object-aware knowledge extraction to update the parameters in both the backbone and RPN.

\subsection{Main Performance}

We compare our method against motion-guided methods~\cite{zhu2017flow,feichtenhofer2017detect, wang2018fully}, SELSA~\cite{wu2019sequence}, RDN~\cite{deng2019relation} and MEGA~\cite{chen2020memory}. Expanding from motion-guided methods, FEVOD~\cite{jiang2019learning} directly learns and predicts sampling positions to improve performance. For SELSA~\cite{wu2019sequence}, RDN~\cite{deng2019relation} and MEGA~\cite{chen2020memory}, they all treat the attention operators as core components to improve their performance. Especially the previous state-of-the-art method, MEGA, exploring the idea to empower predictions with context from longer content both globally and locally.
Table~\ref{tab:performance} shows the performance comparison with these state-of-the-art approaches. 
With different backbones and post-processing algorithms, we fully compare our approach with others. Obviously, no matter end-to-end models or performance enhanced with post-processing, our approach consistently achieves the best performances on different backbones.
\vspace{-0.3cm}
\paragraph{End-to-End models.} For fair comparisons, we use naive NMS as the post-processing operation when evaluating different end-to-end models. 
With the ResNet-101 backbone, our approach can achieve $83.93\%$ mAP, with $1.33\%$ absolute improvement over the baseline. Among all competitors, our approach gains at least $1.03\%$ improvement. 
As one of the most similar methods, MEGA
benefits $1.0\%$ from the global context with the Long Range Memory~(LRM) module. Although it can get access to more temporal contexts in larger temporal windows, the recurrent updated LRM is not able to get an overview of the input video. Hence, our approach is much more flexible to process variable-length sequences and brings more improvement. Even with stronger ResNeXt-101, the baseline model still gains an improvement of $1.79\%$ and achieves the new state-of-the-art performance of $86.09\%$ mAP.
\vspace{-0.3cm}
\paragraph{Add post-processing.} As most of the state-of-the-art video object detection approaches could benefit from sequence post-processing, we also compare our approach with their best performances achieved with different post-processing strategies. Instead of BLR, we adopt our Sequence DIoU NMS without extra inputs. Table~\ref{tab:performance} summarizes the results of state-of-the-art methods with different post-processing. Obviously, our approach still performs the best with $85.07\%$ and $86.88\%$ mAP with backbone ResNet-101 and ResNeXt-101, respectively.

\subsection{Ablation Study}

To study the impacts of components in our approach, extensive ablation experiments are conducted. All these experiments mainly start from baseline with the same ResNet-101 backbone.
\vspace{-0.3cm}
\paragraph{Effect of the object-aware knowledge of proposals.} To explore the effect of the object-aware knowledge of proposals in our approach, we show the performance in Table~\ref{tab:ablation1}. 
In addition, we introduce two different strategies to extract object-aware knowledge. Inspired by global context extraction in semantic segmentation~\cite{liu2015parsenet, zhao2017pyramid}, we compute the global context of frames~($gcf$) by averaging the value of feature maps.
We also calculate the mean value of all proposals~($gcp$) to validate the necessary to weight proposals.

Comparing different object-aware knowledge extraction strategies, the global context obtained by $gcf$ even harms performance. It is proven that too much background information in the global context harms the recognition of objects. 
On the positive side, the object-aware knowledge obtained by $gcp$ and our \modulenameshort{} module~($ofa$) can improve performance. It is because proposals mainly focus on the part of the scene where objects may exist and most of the background is filtered out. Especially, the strategy of highlighting proposals with high objectness scores significantly surpasses the baseline by $1.3\%$ mAP. 
With motion-specific mAP, we find that our approach improves the recognition of objects with different motion speeds. Especially for objects with small or medium motion, since more clear object-aware knowledge can be produced, larger benefits are obtained.
Meanwhile, as shown in the ablation experiments of feature splitting~(SP), parallel feature augmentation does not make a significant impact on the performance. Nevertheless, the individual semantic path indeed is a necessary key to combine object-aware knowledge to boost recognition ability.

Additionally, with stronger backbone ResNeXt-101, the absolute improvement is further increased. It is inferred that a more powerful object-aware knowledge is extracted from stronger encoded features and more accurate proposals.

\begin{table}[t]
    \begin{center}
    \begin{tabular}{c c| cccc}
        \hline
            ~OA~&~SP~&~mAP~&~${\rm mAP_s}$~&~${\rm mAP_m}$~&~${\rm mAP_f}$~\\
        \hline
            -~&~-~&~82.69~&~88.00~&~81.35~&~67.10~\\
            -~&~\checkmark~&~82.79~&~88.46~&~81.50~&~66.31~\\
            $gcf$~&~-~&~82.24~&~89.22~&~80.59~&~65.17~\\
            $gcp$~&~\checkmark~&~83.14~&~87.54~&~82.23~&~67.13~\\
            $ofa$~&~-~&~83.29~&~88.30~&~82.05&~67.74~\\
            $ofa$~&~\checkmark~&~\textbf{83.93}~&~\textbf{89.44}~&~\textbf{82.67}~&~\textbf{67.36}~\\
        \hline
    \end{tabular}
    \end{center}
    \caption{Ablation study on the components of the \modulenameshort{} module. Here, we mainly verify the impacts of feature splitting~(SP) and different object-aware knowledge~(OA) extraction strategies. ${\rm mAP_s}$, ${\rm mAP_m}$, ${\rm mAP_f}$ represent mAP(small), mAP(medium), mAP(fast), respectively.} 
    \label{tab:ablation1}
\end{table}

\begin{table}[t]
    \begin{center}
    \begin{tabular}{c |c| c}
        \hline
            ~Network~&~Post-Processing~&~mAP~\\
        \hline
            Baseline~&~NMS~&~82.69~\\
            Baseline~&~Seq DIoU NMS~&~84.20~\\
            RDN~&~NMS~&~81.80~\\
            RDN~&~BLR~&~83.80~\\
            MEGA~&~NMS~&~82.90~\\
            MEGA~&~BLR~&~84.50~\\
        \hline
            Ours~&~NMS~&~83.93~\\
            Ours~&~DIoU NMS~&~84.66~\\
            Ours~&~Seq-NMS~&~84.00~\\
            \textbf{Ours}~&~\textbf{Seq DIoU NMS}~&~\textbf{85.07}~\\
        \hline
    \end{tabular}
    \end{center}
    \caption{Performance comparison with state-of-the-art video object detection models with post-processing methods (e.g. Seq-NMS, BLR, and our Sequence DIoU NMS~(Seq DIoU NMS)).} 
    \label{tab:ablation2}
\end{table}
\vspace{-0.3cm}
\paragraph{Effect of post-processing.} To explore different post-processing methods, we show their performance in Table~\ref{tab:ablation2}. Compared with the common NMS, different sequence post-processing all have greater improvements in the results. At present, the most effective post-processing BLR improved two state-of-the-art networks, MEGA~\cite{chen2020memory} and RDN~\cite{deng2019relation}, by a margin from $1.6\%$ to $2.0\%$.  It can be seen that the improvements of post-processing methods are compressed in more powerful networks. 
Compared with others, our Sequence DIoU NMS improves the baseline by $1.51\%$. Even on our network, of which performance is much higher than other networks, the Sequence DIoU NMS also brings $0.94\%$ improvement. 
To analyze the improvement of Sequence DIoU NMS, we tested the effects of DIoU NMS and Seq-NMS in our approach. 
Sequence DIoU NMS significantly improves the performance of our approach with different backbones, which is more than the straightforward combination of DIoU NMS and Seq-NMS.
\section{Conclusion}
\label{Conclusion}
In this paper, we have presented the \modulename{}~(\modulenameshort{}) module to extract video-level object-aware knowledge of proposals for video object detection. Our \modulenameshort{} module contains two separable parallel paths, \ie, \textit{semantic path} and \textit{localization path} for classification and regression, respectively. In fact, the \modulenameshort{} module improves the performance via incorporating the prior knowledge with well-established pair-wise contexts, which is compatible with any attention-based video object detection methods. Sequence DIoU NMS further boosts the performance at the post-processing stage. 
Extensive experiments on the ImageNet VID dataset have demonstrated the effectiveness of the proposed method.
Future research may focus on introducing video-level object-aware knowledge in other proposal-based vision tasks like object tracking and action recognition.

\bibliography{citation}

\begin{thebibliography}{54}
\providecommand{\natexlab}[1]{#1}
\providecommand{\url}[1]{\texttt{#1}}
\providecommand{\urlprefix}{URL }
\expandafter\ifx\csname urlstyle\endcsname\relax
  \providecommand{\doi}[1]{doi:\discretionary{}{}{}#1}\else
  \providecommand{\doi}{doi:\discretionary{}{}{}\begingroup
  \urlstyle{rm}\Url}\fi

\bibitem[{Bertasius, Torresani, and Shi(2018)}]{bertasius2018object}
Bertasius, G.; Torresani, L.; and Shi, J. 2018.
\newblock Object detection in video with spatiotemporal sampling networks.
\newblock In \emph{Proceedings of the European Conference on Computer Vision
  (ECCV)}, 331--346.

\bibitem[{Cao et~al.(2019)Cao, Xu, Lin, Wei, and Hu}]{cao2019gcnet}
Cao, Y.; Xu, J.; Lin, S.; Wei, F.; and Hu, H. 2019.
\newblock Gcnet: Non-local networks meet squeeze-excitation networks and
  beyond.
\newblock In \emph{Proceedings of the IEEE International Conference on Computer
  Vision Workshops}, 0--0.

\bibitem[{Chen et~al.(2018)Chen, Wang, Yang, Zhang, Xiong, Change~Loy, and
  Lin}]{chen2018optimizing}
Chen, K.; Wang, J.; Yang, S.; Zhang, X.; Xiong, Y.; Change~Loy, C.; and Lin, D.
  2018.
\newblock Optimizing video object detection via a scale-time lattice.
\newblock In \emph{Proceedings of the IEEE conference on computer vision and
  pattern recognition}, 7814--7823.

\bibitem[{Chen et~al.(2020)Chen, Cao, Hu, and Wang}]{chen2020memory}
Chen, Y.; Cao, Y.; Hu, H.; and Wang, L. 2020.
\newblock Memory Enhanced Global-Local Aggregation for Video Object Detection.
\newblock In \emph{Proceedings of the IEEE/CVF Conference on Computer Vision
  and Pattern Recognition}, 10337--10346.

\bibitem[{Dai et~al.(2016)Dai, Li, He, and Sun}]{dai2016r}
Dai, J.; Li, Y.; He, K.; and Sun, J. 2016.
\newblock R-fcn: Object detection via region-based fully convolutional
  networks.
\newblock In \emph{Advances in neural information processing systems},
  379--387.

\bibitem[{Deng et~al.(2019)Deng, Pan, Yao, Zhou, Li, and
  Mei}]{deng2019relation}
Deng, J.; Pan, Y.; Yao, T.; Zhou, W.; Li, H.; and Mei, T. 2019.
\newblock Relation distillation networks for video object detection.
\newblock In \emph{Proceedings of the IEEE International Conference on Computer
  Vision}, 7023--7032.

\bibitem[{Feichtenhofer, Pinz, and Zisserman(2017)}]{feichtenhofer2017detect}
Feichtenhofer, C.; Pinz, A.; and Zisserman, A. 2017.
\newblock Detect to track and track to detect.
\newblock In \emph{Proceedings of the IEEE International Conference on Computer
  Vision}, 3038--3046.

\bibitem[{Fu et~al.(2017)Fu, Liu, Ranga, Tyagi, and Berg}]{fu2017dssd}
Fu, C.-Y.; Liu, W.; Ranga, A.; Tyagi, A.; and Berg, A.~C. 2017.
\newblock Dssd: Deconvolutional single shot detector.
\newblock \emph{arXiv preprint arXiv:1701.06659} .

\bibitem[{Girshick(2015)}]{girshick2015fast}
Girshick, R. 2015.
\newblock Fast r-cnn.
\newblock In \emph{Proceedings of the IEEE international conference on computer
  vision}, 1440--1448.

\bibitem[{Girshick et~al.(2014)Girshick, Donahue, Darrell, and
  Malik}]{girshick2014rich}
Girshick, R.; Donahue, J.; Darrell, T.; and Malik, J. 2014.
\newblock Rich feature hierarchies for accurate object detection and semantic
  segmentation.
\newblock In \emph{Proceedings of the IEEE conference on computer vision and
  pattern recognition}, 580--587.

\bibitem[{Gkioxari and Malik(2015)}]{gkioxari2015finding}
Gkioxari, G.; and Malik, J. 2015.
\newblock Finding action tubes.
\newblock In \emph{Proceedings of the IEEE conference on computer vision and
  pattern recognition}, 759--768.

\bibitem[{Han et~al.(2016)Han, Khorrami, Paine, Ramachandran, Babaeizadeh, Shi,
  Li, Yan, and Huang}]{han2016seq}
Han, W.; Khorrami, P.; Paine, T.~L.; Ramachandran, P.; Babaeizadeh, M.; Shi,
  H.; Li, J.; Yan, S.; and Huang, T.~S. 2016.
\newblock Seq-nms for video object detection.
\newblock \emph{arXiv preprint arXiv:1602.08465} .

\bibitem[{He et~al.(2017)He, Gkioxari, Doll{\'a}r, and Girshick}]{he2017mask}
He, K.; Gkioxari, G.; Doll{\'a}r, P.; and Girshick, R. 2017.
\newblock Mask r-cnn.
\newblock In \emph{Proceedings of the IEEE international conference on computer
  vision}, 2961--2969.

\bibitem[{He et~al.(2015)He, Zhang, Ren, and Sun}]{he2015spatial}
He, K.; Zhang, X.; Ren, S.; and Sun, J. 2015.
\newblock Spatial pyramid pooling in deep convolutional networks for visual
  recognition.
\newblock \emph{IEEE transactions on pattern analysis and machine intelligence}
  37(9): 1904--1916.

\bibitem[{He et~al.(2016)He, Zhang, Ren, and Sun}]{he2016deep}
He, K.; Zhang, X.; Ren, S.; and Sun, J. 2016.
\newblock Deep residual learning for image recognition.
\newblock In \emph{Proceedings of the IEEE conference on computer vision and
  pattern recognition}, 770--778.

\bibitem[{Hu et~al.(2018)Hu, Gu, Zhang, Dai, and Wei}]{hu2018relation}
Hu, H.; Gu, J.; Zhang, Z.; Dai, J.; and Wei, Y. 2018.
\newblock Relation networks for object detection.
\newblock In \emph{Proceedings of the IEEE Conference on Computer Vision and
  Pattern Recognition}, 3588--3597.

\bibitem[{Huang et~al.(2017)Huang, Liu, Van Der~Maaten, and
  Weinberger}]{huang2017densely}
Huang, G.; Liu, Z.; Van Der~Maaten, L.; and Weinberger, K.~Q. 2017.
\newblock Densely connected convolutional networks.
\newblock In \emph{Proceedings of the IEEE conference on computer vision and
  pattern recognition}, 4700--4708.

\bibitem[{Jiang et~al.(2019)Jiang, Liu, Yang, Liu, Gao, Zhang, Xiang, and
  Pan}]{jiang2019learning}
Jiang, Z.; Liu, Y.; Yang, C.; Liu, J.; Gao, P.; Zhang, Q.; Xiang, S.; and Pan,
  C. 2019.
\newblock Learning Where to Focus for Efficient Video Object Detection.

\bibitem[{Kang et~al.(2017{\natexlab{a}})Kang, Li, Xiao, Ouyang, Yan, Liu, and
  Wang}]{kang2017object}
Kang, K.; Li, H.; Xiao, T.; Ouyang, W.; Yan, J.; Liu, X.; and Wang, X.
  2017{\natexlab{a}}.
\newblock Object detection in videos with tubelet proposal networks.
\newblock In \emph{Proceedings of the IEEE Conference on Computer Vision and
  Pattern Recognition}, 727--735.

\bibitem[{Kang et~al.(2017{\natexlab{b}})Kang, Li, Yan, Zeng, Yang, Xiao,
  Zhang, Wang, Wang, Wang et~al.}]{kang2017t}
Kang, K.; Li, H.; Yan, J.; Zeng, X.; Yang, B.; Xiao, T.; Zhang, C.; Wang, Z.;
  Wang, R.; Wang, X.; et~al. 2017{\natexlab{b}}.
\newblock T-cnn: Tubelets with convolutional neural networks for object
  detection from videos.
\newblock \emph{IEEE Transactions on Circuits and Systems for Video Technology}
  28(10): 2896--2907.

\bibitem[{Kang et~al.(2016)Kang, Ouyang, Li, and Wang}]{kang2016object}
Kang, K.; Ouyang, W.; Li, H.; and Wang, X. 2016.
\newblock Object detection from video tubelets with convolutional neural
  networks.
\newblock In \emph{Proceedings of the IEEE conference on computer vision and
  pattern recognition}, 817--825.

\bibitem[{Krizhevsky, Sutskever, and Hinton(2012)}]{krizhevsky2012imagenet}
Krizhevsky, A.; Sutskever, I.; and Hinton, G.~E. 2012.
\newblock Imagenet classification with deep convolutional neural networks.
\newblock In \emph{Advances in neural information processing systems},
  1097--1105.

\bibitem[{Law and Deng(2018)}]{law2018cornernet}
Law, H.; and Deng, J. 2018.
\newblock Cornernet: Detecting objects as paired keypoints.
\newblock In \emph{Proceedings of the European Conference on Computer Vision
  (ECCV)}, 734--750.

\bibitem[{Li et~al.(2018{\natexlab{a}})Li, Qiu, Dai, Yao, and
  Mei}]{li2018recurrent}
Li, D.; Qiu, Z.; Dai, Q.; Yao, T.; and Mei, T. 2018{\natexlab{a}}.
\newblock Recurrent tubelet proposal and recognition networks for action
  detection.
\newblock In \emph{Proceedings of the European conference on computer vision
  (ECCV)}, 303--318.

\bibitem[{Li et~al.(2018{\natexlab{b}})Li, Yao, Duan, Mei, and
  Rui}]{li2018unified}
Li, D.; Yao, T.; Duan, L.-Y.; Mei, T.; and Rui, Y. 2018{\natexlab{b}}.
\newblock Unified spatio-temporal attention networks for action recognition in
  videos.
\newblock \emph{IEEE Transactions on Multimedia} 21(2): 416--428.

\bibitem[{Lin et~al.(2017{\natexlab{a}})Lin, Doll{\'a}r, Girshick, He,
  Hariharan, and Belongie}]{lin2017feature}
Lin, T.-Y.; Doll{\'a}r, P.; Girshick, R.; He, K.; Hariharan, B.; and Belongie,
  S. 2017{\natexlab{a}}.
\newblock Feature pyramid networks for object detection.
\newblock In \emph{Proceedings of the IEEE conference on computer vision and
  pattern recognition}, 2117--2125.

\bibitem[{Lin et~al.(2017{\natexlab{b}})Lin, Goyal, Girshick, He, and
  Doll{\'a}r}]{lin2017focal}
Lin, T.-Y.; Goyal, P.; Girshick, R.; He, K.; and Doll{\'a}r, P.
  2017{\natexlab{b}}.
\newblock Focal loss for dense object detection.
\newblock In \emph{Proceedings of the IEEE international conference on computer
  vision}, 2980--2988.

\bibitem[{Lin et~al.(2014)Lin, Maire, Belongie, Hays, Perona, Ramanan,
  Doll{\'a}r, and Zitnick}]{lin2014microsoft}
Lin, T.-Y.; Maire, M.; Belongie, S.; Hays, J.; Perona, P.; Ramanan, D.;
  Doll{\'a}r, P.; and Zitnick, C.~L. 2014.
\newblock Microsoft coco: Common objects in context.
\newblock In \emph{European conference on computer vision}, 740--755. Springer.

\bibitem[{Liu et~al.(2016)Liu, Anguelov, Erhan, Szegedy, Reed, Fu, and
  Berg}]{liu2016ssd}
Liu, W.; Anguelov, D.; Erhan, D.; Szegedy, C.; Reed, S.; Fu, C.-Y.; and Berg,
  A.~C. 2016.
\newblock Ssd: Single shot multibox detector.
\newblock In \emph{European conference on computer vision}, 21--37. Springer.

\bibitem[{Liu, Rabinovich, and Berg(2015)}]{liu2015parsenet}
Liu, W.; Rabinovich, A.; and Berg, A.~C. 2015.
\newblock Parsenet: Looking wider to see better.
\newblock \emph{arXiv preprint arXiv:1506.04579} .

\bibitem[{Pan et~al.(2016)Pan, Li, Yao, Mei, Li, and Rui}]{pan2016learning}
Pan, Y.; Li, Y.; Yao, T.; Mei, T.; Li, H.; and Rui, Y. 2016.
\newblock Learning Deep Intrinsic Video Representation by Exploring Temporal
  Coherence and Graph Structure.
\newblock In \emph{IJCAI}, 3832--3838.

\bibitem[{Qiu, Yao, and Mei(2017)}]{qiu2017learning}
Qiu, Z.; Yao, T.; and Mei, T. 2017.
\newblock Learning spatio-temporal representation with pseudo-3d residual
  networks.
\newblock In \emph{proceedings of the IEEE International Conference on Computer
  Vision}, 5533--5541.

\bibitem[{Redmon et~al.(2016)Redmon, Divvala, Girshick, and
  Farhadi}]{redmon2016you}
Redmon, J.; Divvala, S.; Girshick, R.; and Farhadi, A. 2016.
\newblock You only look once: Unified, real-time object detection.
\newblock In \emph{Proceedings of the IEEE conference on computer vision and
  pattern recognition}, 779--788.

\bibitem[{Redmon and Farhadi(2017)}]{redmon2017yolo9000}
Redmon, J.; and Farhadi, A. 2017.
\newblock YOLO9000: better, faster, stronger.
\newblock In \emph{Proceedings of the IEEE conference on computer vision and
  pattern recognition}, 7263--7271.

\bibitem[{Redmon and Farhadi(2018)}]{redmon2018yolov3}
Redmon, J.; and Farhadi, A. 2018.
\newblock Yolov3: An incremental improvement.
\newblock \emph{arXiv preprint arXiv:1804.02767} .

\bibitem[{Ren et~al.(2015)Ren, He, Girshick, and Sun}]{ren2015faster}
Ren, S.; He, K.; Girshick, R.; and Sun, J. 2015.
\newblock Faster r-cnn: Towards real-time object detection with region proposal
  networks.
\newblock In \emph{Advances in neural information processing systems}, 91--99.

\bibitem[{Russakovsky et~al.(2015)Russakovsky, Deng, Su, Krause, Satheesh, Ma,
  Huang, Karpathy, Khosla, Bernstein et~al.}]{russakovsky2015imagenet}
Russakovsky, O.; Deng, J.; Su, H.; Krause, J.; Satheesh, S.; Ma, S.; Huang, Z.;
  Karpathy, A.; Khosla, A.; Bernstein, M.; et~al. 2015.
\newblock Imagenet large scale visual recognition challenge.
\newblock \emph{International journal of computer vision} 115(3): 211--252.

\bibitem[{Simonyan and Zisserman(2014{\natexlab{a}})}]{simonyan2014two}
Simonyan, K.; and Zisserman, A. 2014{\natexlab{a}}.
\newblock Two-stream convolutional networks for action recognition in videos.
\newblock In \emph{Advances in neural information processing systems},
  568--576.

\bibitem[{Simonyan and Zisserman(2014{\natexlab{b}})}]{simonyan2014very}
Simonyan, K.; and Zisserman, A. 2014{\natexlab{b}}.
\newblock Very deep convolutional networks for large-scale image recognition.
\newblock \emph{arXiv preprint arXiv:1409.1556} .

\bibitem[{Szegedy et~al.(2015)Szegedy, Liu, Jia, Sermanet, Reed, Anguelov,
  Erhan, Vanhoucke, and Rabinovich}]{szegedy2015going}
Szegedy, C.; Liu, W.; Jia, Y.; Sermanet, P.; Reed, S.; Anguelov, D.; Erhan, D.;
  Vanhoucke, V.; and Rabinovich, A. 2015.
\newblock Going deeper with convolutions.
\newblock In \emph{Proceedings of the IEEE conference on computer vision and
  pattern recognition}, 1--9.

\bibitem[{Vaswani et~al.(2017)Vaswani, Shazeer, Parmar, Uszkoreit, Jones,
  Gomez, Kaiser, and Polosukhin}]{vaswani2017attention}
Vaswani, A.; Shazeer, N.; Parmar, N.; Uszkoreit, J.; Jones, L.; Gomez, A.~N.;
  Kaiser, {\L}.; and Polosukhin, I. 2017.
\newblock Attention is all you need.
\newblock In \emph{Advances in neural information processing systems},
  5998--6008.

\bibitem[{Wang et~al.(2018)Wang, Zhou, Yan, and Deng}]{wang2018fully}
Wang, S.; Zhou, Y.; Yan, J.; and Deng, Z. 2018.
\newblock Fully motion-aware network for video object detection.
\newblock In \emph{Proceedings of the European Conference on Computer Vision
  (ECCV)}, 542--557.

\bibitem[{Wang and Gupta(2018)}]{wang2018videos}
Wang, X.; and Gupta, A. 2018.
\newblock Videos as space-time region graphs.
\newblock In \emph{Proceedings of the European conference on computer vision
  (ECCV)}, 399--417.

\bibitem[{Wu et~al.(2019)Wu, Chen, Wang, and Zhang}]{wu2019sequence}
Wu, H.; Chen, Y.; Wang, N.; and Zhang, Z. 2019.
\newblock Sequence level semantics aggregation for video object detection.
\newblock In \emph{Proceedings of the IEEE International Conference on Computer
  Vision}, 9217--9225.

\bibitem[{Wu et~al.(2020)Wu, Chen, Yuan, Liu, Wang, Li, and
  Fu}]{wu2020rethinking}
Wu, Y.; Chen, Y.; Yuan, L.; Liu, Z.; Wang, L.; Li, H.; and Fu, Y. 2020.
\newblock Rethinking Classification and Localization for Object Detection.
\newblock In \emph{Proceedings of the IEEE/CVF Conference on Computer Vision
  and Pattern Recognition}, 10186--10195.

\bibitem[{Xiao and Jae~Lee(2018)}]{xiao2018video}
Xiao, F.; and Jae~Lee, Y. 2018.
\newblock Video object detection with an aligned spatial-temporal memory.
\newblock In \emph{Proceedings of the European Conference on Computer Vision
  (ECCV)}, 485--501.

\bibitem[{Xie et~al.(2017)Xie, Girshick, Doll{\'a}r, Tu, and
  He}]{xie2017aggregated}
Xie, S.; Girshick, R.; Doll{\'a}r, P.; Tu, Z.; and He, K. 2017.
\newblock Aggregated residual transformations for deep neural networks.
\newblock In \emph{Proceedings of the IEEE conference on computer vision and
  pattern recognition}, 1492--1500.

\bibitem[{Zhang et~al.(2018)Zhang, Dana, Shi, Zhang, Wang, Tyagi, and
  Agrawal}]{zhang2018context}
Zhang, H.; Dana, K.; Shi, J.; Zhang, Z.; Wang, X.; Tyagi, A.; and Agrawal, A.
  2018.
\newblock Context encoding for semantic segmentation.
\newblock In \emph{Proceedings of the IEEE conference on Computer Vision and
  Pattern Recognition}, 7151--7160.

\bibitem[{Zhao et~al.(2017)Zhao, Shi, Qi, Wang, and Jia}]{zhao2017pyramid}
Zhao, H.; Shi, J.; Qi, X.; Wang, X.; and Jia, J. 2017.
\newblock Pyramid scene parsing network.
\newblock In \emph{Proceedings of the IEEE conference on computer vision and
  pattern recognition}, 2881--2890.

\bibitem[{Zheng et~al.(2020)Zheng, Wang, Liu, Li, Ye, and
  Ren}]{zheng2020distance}
Zheng, Z.; Wang, P.; Liu, W.; Li, J.; Ye, R.; and Ren, D. 2020.
\newblock Distance-IoU Loss: Faster and Better Learning for Bounding Box
  Regression.
\newblock In \emph{AAAI}, 12993--13000.

\bibitem[{Zhou, Wang, and Kr{\"a}henb{\"u}hl(2019)}]{zhou2019objects}
Zhou, X.; Wang, D.; and Kr{\"a}henb{\"u}hl, P. 2019.
\newblock Objects as points.
\newblock \emph{arXiv preprint arXiv:1904.07850} .

\bibitem[{Zhu et~al.(2018)Zhu, Dai, Yuan, and Wei}]{zhu2018towards}
Zhu, X.; Dai, J.; Yuan, L.; and Wei, Y. 2018.
\newblock Towards high performance video object detection.
\newblock In \emph{Proceedings of the IEEE Conference on Computer Vision and
  Pattern Recognition}, 7210--7218.

\bibitem[{Zhu et~al.(2017{\natexlab{a}})Zhu, Wang, Dai, Yuan, and
  Wei}]{zhu2017flow}
Zhu, X.; Wang, Y.; Dai, J.; Yuan, L.; and Wei, Y. 2017{\natexlab{a}}.
\newblock Flow-guided feature aggregation for video object detection.
\newblock In \emph{Proceedings of the IEEE International Conference on Computer
  Vision}, 408--417.

\bibitem[{Zhu et~al.(2017{\natexlab{b}})Zhu, Xiong, Dai, Yuan, and
  Wei}]{zhu2017deep}
Zhu, X.; Xiong, Y.; Dai, J.; Yuan, L.; and Wei, Y. 2017{\natexlab{b}}.
\newblock Deep feature flow for video recognition.
\newblock In \emph{Proceedings of the IEEE conference on computer vision and
  pattern recognition}, 2349--2358.

\end{thebibliography}
\end{document}